\documentclass[11pt]{article}

\usepackage{acl}

\usepackage{times}
\usepackage{latexsym}

\usepackage[T1]{fontenc}

\usepackage[utf8]{inputenc}

\usepackage{microtype}

\usepackage{inconsolata}

\usepackage{graphicx}
\usepackage{amsmath}
\usepackage{amssymb}
\usepackage{booktabs}
\usepackage{multirow}
\usepackage{algorithm}
\usepackage{algpseudocode}
\usepackage[most]{tcolorbox}
\definecolor{promptcardframe}{HTML}{8EA9D6}
\definecolor{promptcardtitle}{HTML}{EAF2FF}
\definecolor{promptcardbody}{HTML}{FBFCFF}
\newtcolorbox{promptcard}[1]{%
  enhanced,
  breakable,
  colback=promptcardbody,
  colframe=promptcardframe,
  colbacktitle=promptcardtitle,
  coltitle=black,
  title={#1},
  fonttitle=\small\bfseries,
  fontupper=\small,
  boxrule=0.55pt,
  arc=1.5mm,
  left=1.2mm,
  right=1.2mm,
  top=0.9mm,
  bottom=0.9mm,
  toptitle=0.75mm,
  bottomtitle=0.75mm,
  before skip=0.75\baselineskip,
  after skip=0.75\baselineskip
}
\title{Stop Training Against Soft Users: Mentalized Adversarial Self-Play for Proactive Dialogue}

\author{
  \textbf{Zanwu Liu\textsuperscript{2,*}},
  \textbf{Dan Zhang\textsuperscript{1,*}},
  \textbf{Xiaodi Wang\textsuperscript{1,*}},
  \textbf{Yongbo Mao\textsuperscript{1}},
  \textbf{Xiaoyang Guo\textsuperscript{1}}
  \\
  \textbf{Yonglin Wu\textsuperscript{1}},
  \textbf{Peiyue Jiang\textsuperscript{1}},
  \textbf{Yangyang Wei\textsuperscript{1}},
  \textbf{Zecheng Lin\textsuperscript{1,\textdagger}}
  \\
  \textsuperscript{1}Dongchezu Technology Co., Ltd., Beijing, China
  \\
  \textsuperscript{2}School of Artificial Intelligence, Beihang University, Beijing, China
  \\
  {\small
    \textsuperscript{*}Equal contribution.
    \textsuperscript{\textdagger}Corresponding author.
  }
  \\
  \small{
    \textbf{First author:} \href{mailto:liuaudi801@gmail.com}{liuaudi801@gmail.com}
    \quad
    \textbf{Correspondence:} \href{mailto:chengyi.2024@dcarlife.com}{chengyi.2024@dcarlife.com}
  }
}

\begin{document}
\maketitle
\begin{abstract}
Proactive dialogue agents must guide multi-turn interactions toward task goals while adapting to users whose intentions and attitudes evolve. Existing planners improve system side action selection through search, learned policies, or user aware planning, but are often trained and evaluated against fixed or overly compliant simulators; as a result, high simulated success can reflect soft user bias rather than robust dialogue ability. We introduce \textsc{MASP}, a \textbf{Mentalized Adversarial Self Play} framework that treats proactive dialogue as a partially observable game between a system and a rationally resisting user. \textsc{MASP} represents the user's evolving mental state with a BDI based latent state, separates reward side teacher mentalization from policy side student mentalization, and alternates self play updates for the system and user policies. This turns robustness to rational resistance into a training objective rather than only an evaluation condition. Across diverse proactive dialogue benchmarks, compared with the widely used SFT and Planner-only approaches, \textsc{MASP} substantially improves success rate (SR) and rationality constrained success (RCSR), reduces average turns (AT), and performs better against hard users who make the task goal more difficult to achieve.
\end{abstract}
\section{Introduction}
\label{sec:introduction}

Proactive dialogue requires more than a fluent next response. In tasks such as persuasion, emotional support, empathy, and negotiation, an agent must guide a multi-turn interaction toward a goal while anticipating how each utterance may change the user's beliefs, desires, and decisions. Although large language models (LLMs) are strong reactive conversationalists, proactive dialogue remains challenging because next-token optimization alone does not provide long-range strategic adaptation \citep{he-etal-2024-planning,He2024SimulationFreeHL}. Recent work has studied various planning and simulation approaches to this challenge \cite{Wang2023TargetorientedPD,Hong2023ZeroShotGD,Liu2025ProactiveEvalAU,Wu2025InterpersonalMM,zhang-etal-2025-enhancing-goal-oriented}.

Most progress, however, remains centered on planners. Prior methods improve system-side action selection through trainable planners, search, and user world modeling \citep{Deng2023PlugandPlayPP,he-etal-2024-planning,zhang-etal-2024-strength,He2025SimulatingBP,He2024SimulationFreeHL,Rakib2025DialogXpertDI,Kim2025PRINCIPLESSS,dao-liao-2025-one,dong-etal-2025-protod}. Despite their different mechanisms, they largely preserve the same training asymmetry: the system is optimized as the sole strategic learner, whereas the user is fixed, externally specified, or only weakly modeled.

This asymmetry can lead to misleading progress. A policy trained against a fixed or overly permissive simulator may achieve high simulated success while learning brittle behavior that fails once the user is less compliant \citep{kreyssig-etal-2018-neural,Keizer2023AdversarialUS,Zhang2024TCRS}. Recent LLM-based work moves beyond prompted role-play toward dedicated user LMs, profile-conditioned simulators, and iterative adversarial refinement \cite{Davidson2023UserSW,Sekulic2024ReliableLU,Naous2025FlippingTD,Zhu2025DIALDI}. These works make the soft user problem explicit, but they mainly strengthen the simulator or evaluation environment; they do not directly answer how a proactive policy should be trained so that its gains survive stronger, more strategic, and partially observed users.

A natural way to address this issue is to model the user's evolving internal state. Recent studies have examined intention modeling, user state tracking, and theory of mind for persuasion and emotional support \cite{Yu2025PersuasiveToMAB,Zhang2025IntentionESCAI,Park2026PsyProbePA}, yet such information remains an auxiliary feature or simulator-side component rather than the basis of the training interaction. We propose \textsc{MASP} (\textbf{M}entalized \textbf{A}dversarial \textbf{S}elf \textbf{P}lay)\footnote{Code and resources are available at: \url{https://github.com/Tecitomous/MASP-Proactive-Dialogue}.}, a training paradigm that instead formulates proactive dialogue as a partially observable game between two agents over a shared latent mental state space. Each turn is governed by a hidden user state represented by free-form \emph{belief}, \emph{desire}, and \emph{intention} (BDI) together with compact affective variables. The user policy conditions on this hidden state, while the system observes only an estimate inferred from dialogue history; consequently, the system is trained not only to choose useful utterances, but also to infer who it is talking to and how that person changes over time.

The key design of \textsc{MASP} is to make mentalization part of the optimization itself. A frozen teacher mentalizer defines the latent state used by reward computation, whereas the system observes only a student estimate, reducing the risk that the policy directly exploits an adaptive reward state estimator. On top of this latent state, \textsc{MASP} defines dense progress toward a task-specific target mind state, measured relative to the user's own starting state. The system is rewarded for both moving the user toward the target and accurately inferring the current state, while the user is rewarded for rational resistance rather than unconditional rejection. This formulation differs from planner-focused approaches by coupling opponent modeling, long-horizon dialogue control, and simulator robustness within a single training game; it also differs from work that only builds simulators because the user model is not merely an evaluation tool, but an adaptive training opponent.

We evaluate \textsc{MASP} on proactive dialogue benchmarks spanning persuasion, emotional support, empathy, and negotiation, with a focus on rationality-constrained success under difficult user simulators. Our contributions are threefold: (1) to our knowledge, \textsc{MASP} is the first multi-turn dialogue training framework to explicitly model BDI mental states and let those states evolve during training; (2) we are the first to apply self-play to a multi-turn dialogue training framework, proposing a mentalized self-play method that jointly trains the system and user and is designed to avoid the soft user problem in practical applications; and (3) to address the fact that widely used evaluation rules do not capture model performance against difficult users in realistic settings, we evaluate models under difficult user simulators and introduce rationality-constrained success rate to better reflect performance with near-realistic users.

\section{Related Work}

Proactive dialogue planning has progressed from search at test time to trainable planners and policy learning with user information. GDP-Zero uses Monte Carlo Tree Search over LLM simulated continuations \citep{Yu2023PromptBasedMT}, while PPDPP trains a planner module with supervised fine tuning and self play reinforcement learning \citep{Deng2023PlugandPlayPP}. Later methods improve planning through dual process deliberation, diversified user simulation, latent policy learning, intrinsic user world modeling, and online selection based on value estimates over frozen LLM proposals \citep{he-etal-2024-planning,zhang-etal-2024-strength,He2024SimulationFreeHL,He2025SimulatingBP,Rakib2025DialogXpertDI}. These methods strengthen system action selection, but still largely optimize a system planner against fixed, prompted, or otherwise externally specified users.

User simulation is central to dialogue training and evaluation, but simulator realism remains a bottleneck \citep{Shi2019HowTB,Davidson2023UserSW,sekulic-etal-2024-reliable,luo-etal-2024-duetsim,wang-etal-2025-know}. Prior work shows that simulator choice affects policy quality and transferability, and that offline policy learning can overfit to simulator artifacts \citep{kreyssig-etal-2018-neural,Keizer2023AdversarialUS,Zhang2024TCRS}. Recent LLM based studies make this issue sharper: \emph{Flipping the Dialogue} trains dedicated User LMs from converted human assistant conversations \citep{Naous2025FlippingTD}, and DIAL improves simulator realism through iterative adversarial refinement \citep{Zhu2025DIALDI}. These works improve simulators or evaluation environments, but do not directly turn user resistance into a coupled training objective for the system policy.

\textsc{MASP} also connects to adversarial and multiagent training. SMAC-Hard shows that narrow opponent distributions can induce brittle multiagent policies \citep{Deng2024SMACHard}, and GAIA formulates noncollaborative dialogue planning as a two player zero sum game with self play and Chain-of-Mind reasoning \citep{wang-etal-2025-battling}. \textsc{MASP} shares the concern that soft or narrow user models produce brittle policies, but differs by modeling proactive dialogue as a partially observable game over a latent mental state geometry: the system sees only a student estimate of the user's hidden state, while the user is trained to resist rationally.

\section{Method}

\label{sec:method}

\begin{figure*}[t]
\centering
\includegraphics[width=\textwidth]{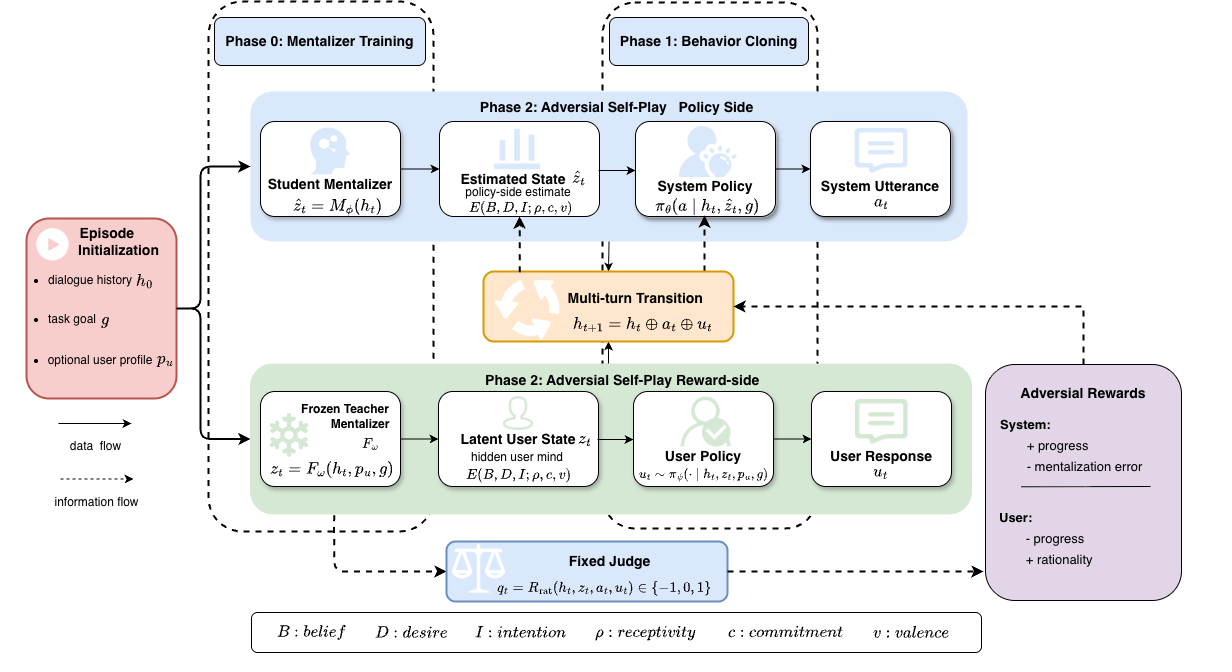}
\caption{
Overview of \textsc{MASP}. The framework trains BDI mentalizers, warm starts system and user policies with behavior cloning, and then performs alternating GRPO self play with latent progress, rationality, and KL regularization to reference policies.
}
\label{fig:masp_framework}
\vspace{-1.5em}
\end{figure*}

As shown in Figure~\ref{fig:masp_framework}, \textsc{MASP} is organized as a three-phase training pipeline. Phase~0 trains BDI mentalizers from silver annotations; Phase~1 uses behavior cloning to obtain warm-start and reference policies for the system and user; Phase~2 alternates GRPO updates for the system and user policies with frozen reward-side modules.

\subsection{Problem Formulation}
\label{subsec:problem}

An episode is defined as
\begin{equation}
\xi = (p_u, g, h_0),
\end{equation}
where $p_u$ denotes an optional user profile, $g$ is the task goal, and $h_0$ is the initial dialogue context. If a dataset does not provide user profiles, we simply set $p_u=\varnothing$.

At turn $t$, the current history prefix is denoted by $h_t$. The system first generates an utterance $a_t$, after which the user generates a response $u_t$, yielding the transition
\begin{equation}
h_{t+1} = h_t \oplus a_t \oplus u_t .
\end{equation}
We model this interaction as a partially observable game between two agents. The user policy conditions on its current latent mental state $z_t$, while the system policy only has access to an estimate $\hat z_t$:
\begin{equation}
a_t \sim \pi_\theta(\cdot \mid h_t, \hat z_t, g),
\end{equation}
\begin{equation}
u_t \sim \pi_\psi(\cdot \mid h_t, z_t, p_u, g).
\end{equation}
Here, $\pi_\theta$ is the system policy and $\pi_\psi$ is the user policy. The asymmetry between $z_t$ and $\hat z_t$ is essential: unlike user aware planners that condition directly on fixed personas or externally provided user descriptions, \textsc{MASP} treats the user's internal state as hidden and dynamically evolving across turns.

The two agents optimize different objectives. Let $r_t^S$ and $r_t^U$ denote the reward at each turn for the system and the user, respectively. Their discounted objectives are
\begin{equation}
\begin{aligned}
J_S(\theta) &= \mathbb{E}\!\left[\sum_{t=0}^{T-1}\gamma^t r_t^S\right], \\
J_U(\psi) &= \mathbb{E}\!\left[\sum_{t=0}^{T-1}\gamma^t r_t^U\right].
\end{aligned}
\end{equation}
All reported self-play runs use $\gamma=0.97$. As we show below, both rewards are defined in a shared latent state geometry, but they induce different behaviors: the system is rewarded for progress and accurate mentalization, whereas the user is rewarded for rational resistance.

\subsection{Latent Mental State and Mentalization}
\label{subsec:latent}

A central question is how to represent the user's internal state. A pure dialogue embedding is flexible but opaque, making it difficult to specify meaningful target states; task-specific slot schemas are interpretable but transfer poorly across domains. BDI provides a useful middle ground: beliefs, desires, and intentions are semantically interpretable enough to anchor target state design, yet general enough to apply across persuasion, support, empathy, and negotiation. We therefore adopt a BDI-based hybrid representation:
\begin{equation}
m_t = (B_t, D_t, I_t, \rho_t, c_t, v_t),
\end{equation}
where $B_t$, $D_t$, and $I_t$ are free form natural language descriptions of the user's \emph{belief}, \emph{desire}, and \emph{intention}, respectively; $\rho_t \in [0,1]$ denotes receptivity, $c_t \in [0,1]$ denotes commitment or confidence, and $v_t \in [-1,1]$ denotes valence. The textual BDI component captures \emph{what} the user currently believes, wants, and plans to do, while the three scalar variables provide a compact affective geometry that remains stable across tasks.

We encode the three textual BDI components separately with a frozen text encoder $E(\cdot)$. In our implementation, $E$ is the corresponding frozen Qwen3 backbone used as a sentence encoder: we mean-pool its final hidden states and $\ell_2$ normalize the pooled vector.
\begin{equation}
e_B=E(B_t),\quad
e_D=E(D_t),\quad
e_I=E(I_t),
\end{equation}
where each component embedding is normalized to unit length. We then define the final vector state as
\begin{equation}
z_t = [e_B;e_D;e_I;\sqrt{\alpha_\rho}\rho_t;\sqrt{\alpha_c}c_t;\sqrt{\alpha_v}v_t].
\end{equation}
This preconditions the scalar dimensions by the same weights used in the latent metric, while keeping the representation deliberately dynamic: it describes \emph{the user's current state}, not a static persona category. In persuasion, support, empathy, or negotiation, what matters is not only who the user is, but what the user currently believes and how that state changes over the course of a dialogue.

To obtain this latent state, \textsc{MASP} uses two mentalization modules. A teacher mentalizer $F_\omega$ predicts the latent state from dialogue history, optional profile, and task goal:
\begin{equation}
z_t = F_\omega(h_t,p_u,g).
\end{equation}
This teacher is pretrained on offline silver annotations and then frozen. Crucially, it is the only state estimator used on the reward side. The system policy never observes $z_t$ directly. Instead, it receives a student estimate
\begin{equation}
\hat z_t = M_\phi(h_t),
\end{equation}
where $M_\phi$ only conditions on the observable dialogue history. This split between teacher and student is one of the key design choices of \textsc{MASP}. Because the reward depends on the frozen teacher state while the policy only sees a student estimate, the policy must learn to infer the user rather than directly exploit the reward model. In other words, mentalization is not just an auxiliary diagnostic module; it becomes part of the optimization problem itself.

The offline silver annotations also define a BDI cache. The cache stores dialogue-prefix records with silver BDI states, and also stores a session-level initial BDI state for each training dialogue. We use the term BDI prior for the non-parametric pool of these initial training states, together with any available profile text. This prior is used to sample initial user states in self play and to render continuous student estimates into text by nearest-neighbor retrieval in the encoded BDI space.

\subsection{User Relative Progress in Latent Space}
\label{subsec:progress}

For each task, we instantiate a target latent state $z^\star$ from a fixed task-level BDI target template, rather than extracting it from test dialogues. The template contains belief, desire, and intention descriptions for a successful end state, plus scalar targets for receptivity, confidence, and valence; it is encoded with the same $E(\cdot)$ and scalar preconditioning as Eq.~(8). For instance, in persuasion the target describes a receptive user who intends to donate, while in emotional support it describes reduced distress and a more constructive intention. Appendix~\ref{app:target_states} lists the concrete templates used in the experiments. Since the BDI embedding may have much higher dimensionality than the scalar variables, we use a weighted metric. The scalar importance weights are applied by pre-scaling the scalar coordinates in Eq.~(8), through the factors $\sqrt{\alpha_\rho}$, $\sqrt{\alpha_c}$, and $\sqrt{\alpha_v}$. The matrix
\begin{equation}
A = \mathrm{diag}\!\left(\frac{1}{3d}I_{3d},1,1,1\right),
\end{equation}
therefore serves to normalize the high-dimensional textual BDI block relative to the three already weighted scalar coordinates, where $d$ is the encoder dimension for one BDI component. Its scalar diagonal entries are set to one because $\alpha_\rho,\alpha_c,\alpha_v$ have already been absorbed into $z_t$. We denote the corresponding norm by
\begin{equation}
\|x\|_A^2 = x^\top A x.
\end{equation}

Minimizing the absolute distance between $z_t$ and $z^\star$ is inappropriate because different users begin from different initial states: a trajectory that is clearly improving for one user may remain far from the global target, while another user may start close without intervention. We therefore define \emph{user-relative progress} as the projection of the current state onto the direction from the user's own initial state $z_0$ to the target $z^\star$:
\begin{equation}
s_t=
\frac{(z_t-z_0)^\top A (z^\star-z_0)}
{(z^\star-z_0)^\top A (z^\star-z_0)+\varepsilon_{\mathrm{prog}}},
\label{eq:progress_main}
\end{equation}
and define the incremental progress by
\begin{equation}
\Delta s_t = s_{t+1}-s_t.
\end{equation}
We set $\varepsilon_{\mathrm{prog}}=10^{-6}$ for numerical stability. This formulation measures whether the dialogue is moving in the \emph{right direction} from the user's own starting point and yields a dense reward signal at every turn, even when task success itself is sparse.

\begin{algorithm}[t]
\small
\caption{\textsc{MASP} training}
\label{alg:masp_training}
\begin{algorithmic}[1]
\Require Dialogue data $\mathcal{D}$, BDI cache, teacher $F_\omega$, student $M_\phi$, system policy $\pi_\theta$, user policy $\pi_\psi$
\State Train teacher $F_\omega$ on silver BDI annotations and freeze it.
\State Train student $M_\phi$ to predict the teacher state from dialogue history.
\State Warm start $\pi_\theta$ and $\pi_\psi$ with supervised learning; save frozen references $\pi_{\theta_0}$ and $\pi_{\psi_0}$.
\For{outer iteration $k=1,\ldots,K$}
    \Statex \textit{System update}
    \State Freeze $\pi_\psi$ and reward side modules.
    \State Roll out dialogues with $\pi_\theta$ and $\pi_\psi$.
    \State Compute $r_S=w_{\mathrm{prog}}\Delta s-w_{\mathrm{inf}}\|\hat z-z\|_A^2$.
    \State Update $\pi_\theta$ with GRPO style grouped advantages and KL to $\pi_{\theta_0}$.
    \Statex \textit{User update}
    \State Freeze $\pi_\theta$ and reward side modules.
    \State Roll out dialogues with the updated $\pi_\theta$ and current $\pi_\psi$.
    \State Compute $r_U=-w_{\mathrm{prog}}\Delta s+w_{\mathrm{rat}}q$.
    \State Update $\pi_\psi$ with GRPO style grouped advantages and KL to $\pi_{\psi_0}$.
    \Statex \textit{Student refresh}
    \State Collect self play prefixes and teacher targets.
    \State Update $M_\phi$ on self play prefixes mixed with real dialogue prefixes.
\EndFor
\end{algorithmic}
\end{algorithm}

\subsection{Adversarial Rewards}
\label{subsec:rewards}

The system should both influence the user and understand the user. We therefore define the system reward as
\begin{equation}
r_t^S
=
w_{\mathrm{prog}}\Delta s_t
-
w_{\mathrm{inf}}\|\hat z_t-z_t\|_A^2.
\label{eq:system_reward_main}
\end{equation}
The first term rewards movement toward the target state, while the second rewards accurate mentalization. The inference term is especially important: it makes the system optimize not only for \emph{what to say}, but also for \emph{how well it understands the person it is talking to}.

The user policy is trained adversarially, but not as an unconstrained adversary. A user that always rejects everything is not realistic, and a user that always agrees is equally uninformative. To prevent both collapse modes, we define
\begin{equation}
r_t^U
=
-w_{\mathrm{prog}}\Delta s_t
+
w_{\mathrm{rat}} q_t,
\label{eq:user_reward_main}
\end{equation}
where $q_t \in \{-1,0,1\}$ is a fixed rationality judge signal indicating whether the user response is rational under the current mental state. The first term rewards resistance to unjustified state change, while the second rewards coherent and rational behavior. Intuitively, the user is not rewarded for rejecting the system blindly; it is rewarded for \emph{resisting when resistance is warranted} and \emph{yielding when change is genuinely justified}. This is what makes the user a useful training opponent rather than merely a harder simulator.

The scalar rollout rewards in Equations~\ref{eq:system_reward_main} and~\ref{eq:user_reward_main} do not include KL penalties. Instead, both policies are regularized by a token level KL term inside the GRPO training objective against their supervised warm start reference policies, as described in Appendix~\ref{app:policy_opt}. This matches the implementation: rewards provide task, mentalization, and rationality feedback, while the trainer applies the KL term to the reference policy.

\subsection{Training}
\label{subsec:training}

Training follows the three stage pipeline in Algorithm~\ref{alg:masp_training}. We first train teacher and student mentalizers from silver BDI annotations, then use behavior cloning to initialize the system and user policies as warm start references, and finally run alternating GRPO self play with frozen reward side modules. The algorithm is the high level flow used in the implementation; additional optimization and prompting details are provided in Appendix~\ref{app:training_details}.

\section{Experiments}
\label{sec:experiments}


\begin{table*}[!t]
\centering
\footnotesize
\setlength{\tabcolsep}{2.8pt}
\renewcommand{\arraystretch}{0.86}
\caption{
Main results on proactive dialogue benchmarks; soft/hard meanings are defined in Evaluation Metrics.
}
\label{tab:main_results}
\begin{tabular*}{\textwidth}{@{\extracolsep{\fill}}lllcccccc@{}}
\toprule
\multirow{2}{*}{Dataset} &
\multirow{2}{*}{Backbone} &
\multirow{2}{*}{Method} &
\multicolumn{3}{c}{Soft User} &
\multicolumn{3}{c}{Hard User} \\
\cmidrule(lr){4-6}
\cmidrule(lr){7-9}
& & &
SR$\uparrow$ & AT$\downarrow$ & RCSR$\uparrow$ &
SR$\uparrow$ & AT$\downarrow$ & RCSR$\uparrow$ \\
\midrule

\multirow{6}{*}{P4G}
& \multirow{3}{*}{Qwen3-4B}
& MASP
& \textbf{0.859} & \textbf{4.08} & \textbf{0.556}
& \textbf{0.788} & \textbf{4.56} & \textbf{0.566} \\
& & SFT
& 0.616 & 6.08 & 0.384
& 0.576 & 6.23 & 0.455 \\
& & Planner-only
& 0.677 & 4.88 & 0.434
& 0.616 & 5.55 & 0.384 \\
\cmidrule(lr){2-9}
& \multirow{3}{*}{Qwen3-0.6B}
& MASP
& \textbf{0.707} & \textbf{5.14} & \textbf{0.465}
& \textbf{0.626} & \textbf{5.76} & \textbf{0.434} \\
& & SFT
& 0.414 & 6.99 & 0.253
& 0.354 & 7.35 & 0.242 \\
& & Planner-only
& 0.424 & 6.32 & 0.232
& 0.364 & 6.84 & 0.202 \\

\midrule

\multirow{6}{*}{ESConv}
& \multirow{3}{*}{Qwen3-4B}
& MASP
& \textbf{0.990} & \textbf{1.96} & \textbf{0.949}
& \textbf{1.000} & \textbf{1.88} & \textbf{0.964} \\
& & SFT
& 0.467 & 7.17 & 0.410
& 0.390 & 7.27 & 0.333 \\
& & Planner-only
& 0.969 & 2.38 & 0.923
& 0.980 & 2.08 & 0.918 \\
\cmidrule(lr){2-9}
& \multirow{3}{*}{Qwen3-0.6B}
& MASP
& \textbf{1.000} & \textbf{2.11} & \textbf{0.949}
& \textbf{1.000} & \textbf{1.65} & \textbf{0.928} \\
& & SFT
& 0.297 & 7.56 & 0.226
& 0.323 & 7.32 & 0.241 \\
& & Planner-only
& 0.954 & 2.77 & 0.897
& 0.990 & 1.84 & 0.903 \\

\midrule

\multirow{6}{*}{Craigslist}
& \multirow{3}{*}{Qwen3-4B}
& MASP
& \textbf{0.689} & \textbf{4.60} & \textbf{0.073}
& \textbf{0.621} & \textbf{5.61} & \textbf{0.259} \\
& & SFT
& 0.646 & 5.38 & 0.045
& 0.558 & 6.28 & 0.234 \\
& & Planner-only
& 0.431 & 5.67 & 0.020
& 0.315 & 6.86 & 0.091 \\
\cmidrule(lr){2-9}
& \multirow{3}{*}{Qwen3-0.6B}
& MASP
& \textbf{0.728} & \textbf{4.24} & 0.030
& \textbf{0.641} & \textbf{5.50} & \textbf{0.211} \\
& & SFT
& 0.726 & 4.84 & \textbf{0.057}
& 0.547 & 6.47 & 0.191 \\
& & Planner-only
& 0.172 & 7.07 & 0.011
& 0.484 & 6.41 & 0.129 \\

\midrule

\multirow{6}{*}{Empathetic}
& \multirow{3}{*}{Qwen3-4B}
& MASP
& \textbf{0.092} & \textbf{7.46} & \textbf{0.059}
& \textbf{0.072} & \textbf{7.57} & \textbf{0.048} \\
& & SFT
& 0.030 & 7.89 & 0.017
& 0.022 & 7.91 & 0.013 \\
& & Planner-only
& 0.042 & 7.75 & 0.027
& 0.037 & 7.78 & 0.024 \\
\cmidrule(lr){2-9}
& \multirow{3}{*}{Qwen3-0.6B}
& MASP
& \textbf{0.063} & \textbf{7.64} & \textbf{0.044}
& \textbf{0.036} & \textbf{7.78} & \textbf{0.024} \\
& & SFT
& 0.013 & 7.95 & 0.007
& 0.002 & 7.99 & 0.001 \\
& & Planner-only
& 0.032 & 7.82 & 0.023
& 0.021 & 7.88 & 0.013 \\

\bottomrule
\end{tabular*}
\end{table*}

\paragraph{Datasets.}
We evaluate on four dialogue benchmarks with multiple turns, spanning proactive and socially grounded interaction: Persuasion for Good (P4G) \citep{Wang2019PersuasionFG}, ESConv \citep{Liu2021TowardsES}, EmpatheticDialogues \citep{rashkin-etal-2019-towards}, and CraigslistBargain \citep{He2018DecouplingSA}. P4G consists of charity persuasion dialogues where a persuader encourages a persuadee to donate, with user level traits and strategy annotations for a subset of conversations. ESConv contains help seeker and supporter conversations grounded in emotional support strategies. EmpatheticDialogues provides emotion grounded open domain conversations for evaluating empathetic response generation. CraigslistBargain contains price negotiations between buyers and sellers over online listings. Together, these datasets cover persuasion, emotional support, empathy, and negotiation, spanning both cooperative and competitive proactive dialogue settings. Appendix~\ref{app:dataset_details} provides dataset statistics and task specific success criteria.

\paragraph{Closed-loop evaluation.}
Each evaluation episode starts from a held-out initial context. At turn $t$, the system generates $a_t$, the user policy generates $u_t$, and both are appended to the history $h_{t+1}$. The rationality judge then reads the full accumulated dialogue rather than an isolated response. This loop continues until success or the eight-exchange limit. Appendix~\ref{app:common_eval_protocol} provides the complete protocol and a concrete rollout.

\paragraph{Rationality-Constrained Success Rate.}
Conventional success rate can overestimate proactive dialogue ability when a user simulator accepts the system's proposal without sufficient justification. We therefore introduce \textbf{RCSR@T}, which counts a dialogue as successful only when the task is completed and the user behavior up to completion remains rational. After each assistant--user exchange, a fixed LLM-based rationality judge receives the task description, the current BDI state $z_t$, the assistant utterance $a_t$, and the user reply $u_t$. The judge answers \texttt{RATIONAL} or \texttt{NOT\_RATIONAL}; we map these labels to a turn score $q_t=+1$ and $q_t=-1$, respectively, and assign $q_t=0$ when the judge output is unparseable or the call fails. If a dialogue first succeeds at turn $\tau_s \le T$, its success-prefix rationality is
\begin{equation}
\bar q_{\tau_s}=\frac{1}{\tau_s}\sum_{t=1}^{\tau_s} q_t .
\end{equation}
The episode-level rationality-constrained success indicator is
\begin{equation}
\mathrm{RCSR@T}
=
\mathbb{I}\!\left[\tau_s \le T \;\wedge\; \bar q_{\tau_s}\ge \tau_q\right],
\end{equation}
and failed dialogues receive zero. We report the mean of this indicator over evaluation dialogues with $\tau_q=0.70$. This threshold requires the successful prefix to be consistently rational while still tolerating occasional unparseable or weak judgments in longer interactions. Unlike SR alone, RCSR filters successes caused by simulator sycophancy or unjustified concessions, and is therefore closer to evaluating whether a model can succeed with near-realistic users, especially difficult users who should resist until given convincing reasons. The exact judge prompt is given in Appendix~\ref{app:common_eval_protocol}.

\paragraph{Evaluation Metrics.}
For each method, we evaluate under soft and hard user conditions, where \emph{soft} denotes the easier simulator for the system under the dataset's task objective and \emph{hard} denotes the more difficult one. For P4G, ESConv, and EmpatheticDialogues, these correspond to the conventional imitation-style and learned adversarial users, respectively; CraigslistBargain reverses this convention because the system is the buyer and the Phase~1 BC seller is harder for buyer-side utility (see Appendix~\ref{app:craigslist_eval_protocol}). We report \textbf{SR@8}, the fraction of dialogues that reach the task-specific success criterion within $T=8$ turns; \textbf{AT}, the average turns to success with failures counted as $T$; and \textbf{RCSR@8}, success under the rationality constraint defined above. Thus, SR and AT measure task effectiveness and efficiency, while RCSR tests whether success reflects rational user behavior rather than degenerate or overly compliant interactions.

\paragraph{Implementation.}
We compare \textsc{MASP} with two controlled baselines under matched backbones, user settings, turn budgets, and success judges within each dataset. \textbf{SFT} uses the supervised dialogue policy without self play. \textbf{Planner-only} uses the same system side warmup and adaptive planning scaffold as \textsc{MASP}, but does not train the adversarial user policy through mentalized self play. \textbf{MASP} is our full method, which further trains the system and user policies with alternating GRPO style self play. Unless otherwise specified, we report results for both Qwen3-4B and Qwen3-0.6B backbones. Dataset specific success definitions, judge prompts, sampling settings, and additional implementation details are provided in Appendix~\ref{app:eval_details}.

\paragraph{Main Results.}
Table~\ref{tab:main_results} shows that \textsc{MASP} consistently improves over both SFT and Planner-only on the completed benchmarks. On P4G, \textsc{MASP} gives the best SR, AT, and RCSR under both soft and hard users for both backbones, indicating that mentalized self play improves not only task completion but also rationality constrained success. On ESConv, \textsc{MASP} reaches near-ceiling SR, but this reflects the prior success convention where \texttt{somewhat\_better} counts as successful rather than requiring full resolution. We therefore treat ESConv mainly as transfer and efficiency evidence and avoid overinterpreting saturated SR. On CraigslistBargain, \textsc{MASP} again leads on SR and AT for both backbones, including the buyer-side hard setting where the Phase 1 BC seller is less favorable to the buyer. The main exception is Qwen3-0.6B soft RCSR, where SFT slightly exceeds \textsc{MASP}; this is a small speed-rationality tradeoff in the most permissive seller setting, not a failure of the overall negotiation behavior. The very low Planner-only 0.6B soft SR is not a table bug: in the buyer role, the fixed planner scaffold can still miss closure against a seller whose concessions do not match its hand-written closing strategy, so the learned buyer policy is needed to exploit the phase 2 seller distribution. On EmpatheticDialogues, absolute SR is lower because success requires a strict resolved outcome, but \textsc{MASP} still yields the best results for soft and hard users across both backbones. A P4G horizontal comparison with prior systems, reported in Appendix~\ref{app:p4g_external}, shows the same trend in absolute comparison.

\subsection{Ablation Study}
\label{subsec:p4g_ablation}

We further isolate the contribution of each \textsc{MASP} component on P4G, following the component analysis style used in prior P4G systems \citep{Deng2023PlugandPlayPP,Rakib2025DialogXpertDI}. Table~\ref{tab:p4g_ablation} reports Qwen3-4B results on the 99 example test split. The \textbf{No GRPO self play} row keeps only the supervised Phase 1 policy, while \textbf{No adversarial self play} uses the Planner-only scaffold without mentalized adversarial self play. The remaining rows ablate the P4G specific inference scaffold by removing the warmup policy, the adaptive close injection, or both.

\begin{table*}[t]
\centering
\footnotesize
\setlength{\tabcolsep}{3pt}
\renewcommand{\arraystretch}{0.92}
\caption{
P4G ablation study on Qwen3-4B.
All rows use the same 99 example test split, $T=8$, and RCSR threshold $\tau=0.70$.
SR and RCSR are higher better; AT is lower better.
}
\label{tab:p4g_ablation}
\begin{tabular*}{\textwidth}{@{\extracolsep{\fill}}llcccccc@{}}
\toprule
\multirow{2}{*}{Variant} &
\multirow{2}{*}{Removed or changed component} &
\multicolumn{3}{c}{Soft User} &
\multicolumn{3}{c}{Hard User} \\
\cmidrule(lr){3-5}
\cmidrule(lr){6-8}
& &
SR$\uparrow$ & AT$\downarrow$ & RCSR$\uparrow$ &
SR$\uparrow$ & AT$\downarrow$ & RCSR$\uparrow$ \\
\midrule
Full MASP
& none
& \textbf{0.859} & 4.08 & \textbf{0.556}
& \textbf{0.788} & 4.56 & \textbf{0.566} \\
No adaptive close
& remove adaptive close injection
& 0.707 & 4.86 & \textbf{0.556}
& 0.707 & 4.95 & 0.495 \\
No warmup
& remove P4G warmup policy
& 0.828 & \textbf{3.83} & 0.283
& 0.727 & \textbf{4.55} & 0.283 \\
No eval scaffold
& remove warmup and adaptive close
& 0.606 & 6.12 & 0.404
& 0.626 & 6.28 & 0.394 \\
No adversarial self play
& Planner-only scaffold
& 0.677 & 4.88 & 0.434
& 0.616 & 5.55 & 0.384 \\
No GRPO self play
& supervised Phase 1 policy only
& 0.616 & 6.08 & 0.384
& 0.576 & 6.23 & 0.455 \\
\bottomrule
\end{tabular*}
\end{table*}

The full model gives the strongest overall result, especially in SR and hard user RCSR. Removing the adaptive close injection reduces SR by 15.2 points under the soft user and 8.1 points under the hard user, while preserving soft user RCSR. Removing the entire evaluation scaffold hurts both SR and AT, and removing self play components leaves large gaps to full \textsc{MASP}; these trends show that the gains come from both mentalized self play training and the P4G specific inference scaffold.

The \textbf{No warmup} row illustrates why RCSR is needed in addition to SR and AT. Without the warmup policy, the model reaches the lowest AT under both user settings, including 3.83 turns for the soft user, and still keeps a relatively high SR. However, its RCSR drops from 0.556 to 0.283 under the soft user and from 0.566 to 0.283 under the hard user. This pattern suggests that the model can learn a shortcut-like behavior that asks for commitment earlier and sometimes obtains a successful endpoint, but does so before the dialogue has provided enough context or user-state change to satisfy the rationality constraint. The warmup component therefore acts less like a speed booster and more like a stabilizer: it helps the system establish the task, surface the user's stance, and make later persuasion attempts count as rationally grounded successes rather than merely fast successes.

\subsection{Human Validation and Statistical Reliability}
\label{subsec:human_validation}

Because the silver BDI states and rationality signal are LLM based, we additionally audit both components with human annotation. Across 160 dialogue prefixes, the three BDI text fields receive mean scores of 4.55--4.62 out of 5, with 91.9--94.4\% scoring at least 4; the scalar fields obtain MAE of 0.071--0.094 and Spearman correlations of 0.79--0.84. On 425 P4G turns, human labels agree with the rationality judge on 91.8\% of cases ($\kappa=0.801$); across 100 dialogues, mean turn agreement is 91.2\% and human--AI rationality scores correlate at $\rho=0.81$. Full protocols and component-level results are reported in Appendix~\ref{app:synthetic_component_audits}.

For the 99-session Qwen3-4B hard-user P4G evaluation, 10,000 paired bootstrap resamples show that \textsc{MASP}'s improvements over Planner-only have 95\% confidence intervals excluding zero for SR, AT, and RCSR. Against SFT, the intervals exclude zero for SR and AT, while the RCSR interval $[-0.0202, 0.2424]$ includes zero. Of the 21 unsuccessful sessions, 20 contain at least three refusal labels; the most frequent overlapping patterns are dialogue breakdowns and repeated pressure. Appendix~\ref{app:bootstrap_intervals} and Appendix~\ref{app:p4g_failure_analysis} provide the complete intervals and failure analysis.

\subsection{Effectiveness Across Dialogue Turns}
\label{subsec:p4g_turn_progress}

The P4G results show that \textsc{MASP} reaches stronger final performance within the full eight turn budget. We next examine when this advantage appears under the rationality-constrained metric, focusing on the hard user setting where resistance is strongest. On the same Qwen3-4B P4G test split, we compute RCSR@t for intermediate turn budgets and retain SR@8 as endpoint task-success context.

\begin{table}[t]
\centering
\footnotesize
\setlength{\tabcolsep}{3pt}
\renewcommand{\arraystretch}{0.95}
\caption{
Effectiveness across dialogue turns.
All rows use Qwen3-4B on the same 99 example P4G test split under the hard user; RCSR uses $\tau_q=0.70$.
}
\label{tab:p4g_turn_progress}
\begin{tabular*}{\linewidth}{@{\extracolsep{\fill}}lcccc@{}}
\toprule
Method & RCSR@3 & RCSR@4 & RCSR@8 & SR@8 \\
\midrule
\textsc{MASP} & \textbf{0.444} & \textbf{0.485} & \textbf{0.566} & \textbf{0.788} \\
Planner-only & 0.212 & 0.273 & 0.384 & 0.616 \\
SFT & 0.071 & 0.141 & 0.455 & 0.576 \\
\bottomrule
\end{tabular*}
\end{table}

\begin{table}[t]
\centering
\footnotesize
\setlength{\tabcolsep}{2.2pt}
\renewcommand{\arraystretch}{0.95}
\caption{
Robustness across P4G user simulators.
}
\label{tab:p4g_user_simulator_robustness}
\begin{tabular}{@{}lcccc@{}}
\toprule
Method & Avg SR & Worst SR & Avg RCSR & Worst RCSR \\
\midrule
\textsc{MASP} & \textbf{0.795} & \textbf{0.737} & \textbf{0.562} & \textbf{0.556} \\
Planner-only & 0.660 & 0.616 & 0.414 & 0.384 \\
SFT & 0.579 & 0.545 & 0.407 & 0.384 \\
\bottomrule
\end{tabular}
\vspace{-2em}
\end{table}

Table~\ref{tab:p4g_turn_progress} shows that the rationality-constrained gains of \textsc{MASP} emerge well before the last turn under the hard user. \textsc{MASP} reaches 0.444 RCSR@3 and 0.485 RCSR@4, while Planner-only reaches 0.212 and 0.273, and SFT reaches only 0.071 and 0.141. Thus the final improvement is not mainly a late recovery effect, nor is it driven only by fast but poorly grounded successes. Instead, mentalized self play appears to help the system identify user state earlier and choose utterances that produce rationally grounded progress with fewer turns against a difficult user.

\subsection{Robustness Across User Simulators}
\label{subsec:p4g_user_simulator_robustness}

We further test whether the gains of \textsc{MASP} depend on a particular user simulator. On the P4G test split, each system is evaluated against an ensemble of three user simulators $\mathcal{U}$. For a method $m$, we define $\mathrm{WorstSR}(m)=\min_{u\in\mathcal{U}}\mathrm{SR}(m,u)$ and $\mathrm{WorstRCSR}(m)=\min_{u\in\mathcal{U}}\mathrm{RCSR}(m,u)$. Table~\ref{tab:p4g_user_simulator_robustness} reports both average and worst case performance across this ensemble.

\textsc{MASP} is strongest under both average and worst-case views. Its worst-case SR remains 0.737 while Planner-only and SFT drop to 0.616 and 0.545; the same pattern holds for RCSR, where \textsc{MASP} keeps 0.556 while both baselines fall to 0.384. This supports the central claim that mentalized adversarial self-play improves robustness rather than only fitting one simulator.

\section{Conclusion}

We presented \textsc{MASP}, a mentalized adversarial self play framework for proactive dialogue. Instead of training a system policy only against soft or fixed user simulators, \textsc{MASP} models dialogue as a partially observable game between two agents over a shared BDI based latent mental state. A frozen teacher mentalizer provides latent state estimates for reward computation, a student mentalizer supplies state estimates for the policy, and alternating GRPO self play trains the system to make task progress while training the user to resist in rational ways. Across completed proactive dialogue benchmarks, \textsc{MASP} improves success, efficiency, and rationality constrained success over SFT and Planner-only baselines. The P4G robustness analysis further shows that these gains are not confined to a single user simulator, supporting the value of treating user resistance as part of the training objective rather than only as an evaluation condition.

\section*{Limitations}

\textsc{MASP} has several limitations. First, training depends on silver mental state annotations, the teacher mentalizer, and the rationality judge, so bias or noise in these components can affect the learned policies. Second, the shared BDI style geometry is designed to transfer across domains, but may miss specialized user states in some tasks. Third, alternating self play with two policies and auxiliary mentalizers is more expensive than single policy planning. Finally, when a dataset does not provide explicit user profiles ($p_u=\varnothing$), the privileged information gap between the teacher and student mentalizers is reduced, so the split mainly regularizes reward side state estimation rather than providing strong profile conditioned hidden information.


\appendix

\section{Additional Experiments}
\label{app:additional_experiments}

\subsection{P4G Horizontal Comparison}
\label{app:p4g_external}

We additionally compare \textsc{MASP} with representative proactive dialogue systems evaluated on P4G in prior work \citep{Rakib2025DialogXpertDI,He2024SimulationFreeHL}. As shown in Table~\ref{tab:p4g_external}, \textsc{MASP} achieves the best success rate and the lowest average turn count. Compared with LDPP, \textsc{MASP} improves SR by 6.4 points and reduces AT by 1.49 turns; compared with UDP, it improves SR by 26.1 points and reduces AT by 3.63 turns.

\begin{table}[t]
\centering
\small
\setlength{\tabcolsep}{4pt}
\renewcommand{\arraystretch}{0.95}
\caption{
P4G comparison with representative proactive dialogue methods.
}
\label{tab:p4g_external}
\begin{tabular}{llcc}
\toprule
Method & Backbone & AT$\downarrow$ & SR$\uparrow$ \\
\midrule
Standard & -- & 8.32 & 0.468 \\
ProCoT & -- & 7.98 & 0.543 \\
ICL-AIF & -- & 8.09 & 0.465 \\
GDP-Zero & -- & 9.12 & 0.328 \\
TRIP & GPT-3.5 & 8.20 & 0.495 \\
PPDPP & Vicuna-13B & 8.19 & 0.463 \\
UDP & GPT-4o mini & 7.71 & 0.598 \\
LDPP & Qwen1.5-1.8B & 5.57 & 0.795 \\
\midrule
\textsc{MASP} (ours) & Qwen3-4B & \textbf{4.08} & \textbf{0.859} \\
\bottomrule
\end{tabular}
\end{table}

\subsection{Learning Dynamics of Self Play}
\label{app:p4g_training_dynamics}

We inspect whether the final gains are a byproduct of the inference scaffold alone or whether Phase 2 self play contributes additional learning. For the P4G Phase 2 run, we track self play rollout performance after each outer iteration and evaluate saved traces with the same success, average turn count, rationality constrained success, and rationality metrics. Table~\ref{tab:p4g_training_dynamics} shows that rollout SR and reward rise across iterations, indicating that the system policy improves under the changing user policy rather than only benefiting from a fixed prompt scaffold.

\begin{table}[t]
\centering
\footnotesize
\setlength{\tabcolsep}{1.8pt}
\renewcommand{\arraystretch}{0.95}
\caption{
Learning dynamics of self play.
Rollout SR and reward are measured during P4G system update rollouts; Eval metrics are recomputed from held out eval traces.
}
\label{tab:p4g_training_dynamics}
\begin{tabular}{@{}ccccccc@{}}
\toprule
Iter. & Roll. SR & Reward & Eval SR & Eval AT & Eval RCSR & MR \\
\midrule
1 & 0.450 & 0.826 & 0.433 & 7.17 & 0.333 & 0.935 \\
2 & 0.550 & 0.946 & 0.533 & 6.00 & 0.200 & 0.867 \\
3 & 0.713 & 1.209 & 0.560 & 6.50 & 0.360 & 0.883 \\
\bottomrule
\end{tabular}
\vspace{0.5em}
\end{table}

\subsection{Human Validation of Synthetic Components}
\label{app:synthetic_component_audits}

Because the mental-state annotations and rationality signal are produced with LLM-based components, we conducted two additional human--AI agreement studies. These audits evaluate the intermediate supervision and reward signals directly, complementing the success-judge audit in Appendix~\ref{app:human_ai_agreement}.

\paragraph{BDI silver-label audit.}
We sampled 160 dialogue prefixes, with 40 examples from each of P4G, ESConv, EmpatheticDialogues, and CraigslistBargain. Human annotators rated the semantic consistency of belief, desire, and intention on a five-point scale, and independently estimated receptivity, confidence, and valence using the same value ranges as \textsc{MASP}. Table~\ref{tab:bdi_human_audit} shows high consistency for the textual fields and strong agreement for the scalar fields.

\begin{table}[t]
\centering
\small
\setlength{\tabcolsep}{4pt}
\renewcommand{\arraystretch}{0.95}
\caption{Human audit of 160 silver BDI records. Text fields report mean consistency and the percentage of scores at least four; scalar fields report MAE and Spearman's $\rho$.}
\label{tab:bdi_human_audit}
\begin{tabular}{@{}lcc@{}}
\toprule
Component & Primary metric & Secondary metric \\
\midrule
Belief & Mean $4.62$ & $\geq 4$: $94.4\%$ \\
Desire & Mean $4.58$ & $\geq 4$: $93.1\%$ \\
Intention & Mean $4.55$ & $\geq 4$: $91.9\%$ \\
Receptivity & MAE $0.071$ & $\rho=0.84$ \\
Confidence & MAE $0.076$ & $\rho=0.82$ \\
Valence & MAE $0.094$ & $\rho=0.79$ \\
\bottomrule
\end{tabular}
\end{table}

\paragraph{Rationality Judge audit.}
We sampled 100 complete P4G trajectories. Human annotators labeled each user response as rational, irrational, or uncertain using the dialogue context and current BDI state, and also provided a dialogue-level confidence score. As reported in Table~\ref{tab:rationality_human_audit}, the judge reaches $91.8\%$ agreement and Cohen's $\kappa=0.801$ over 425 adjudicated turns. The uncertain rate is $4.1\%$ over all 443 turns, and dialogue-level human--AI rationality scores have Spearman correlation $0.81$.

\begin{table}[t]
\centering
\small
\setlength{\tabcolsep}{4pt}
\renewcommand{\arraystretch}{0.95}
\caption{Human audit of the Rationality Judge on 100 P4G trajectories.}
\label{tab:rationality_human_audit}
\begin{tabular}{@{}llrr@{}}
\toprule
Level & Metric & $N$ & Result \\
\midrule
Turn & Agreement & 425 & $91.8\%$ \\
Turn & Cohen's $\kappa$ & 425 & $0.801$ \\
Turn & Uncertain rate & 443 & $4.1\%$ \\
Dialogue & Mean turn agreement & 100 & $91.2\%$ \\
Dialogue & Rationality $\rho$ & 100 & $0.81$ \\
Confidence & Mean score (1--5) & 100 & $4.68$ \\
\bottomrule
\end{tabular}
\end{table}

These results support the consistency of the synthetic components on the audited samples, but they do not establish performance with real users. Evaluation with legally and ethically collected real-user interactions remains future work.

\subsection{Paired Bootstrap Confidence Intervals}
\label{app:bootstrap_intervals}

Using the stored hard-user P4G traces, we computed paired $95\%$ bootstrap confidence intervals over the same 99 held-out sessions with 10,000 resamples. Table~\ref{tab:p4g_bootstrap} reports differences between \textsc{MASP} and each controlled baseline. The intervals exclude zero for SR, AT, and RCSR against Planner-only, and for SR and AT against SFT. The RCSR difference against SFT is positive but its interval includes zero. These intervals measure uncertainty over held-out sessions rather than training variance across random seeds; multi-seed training stability remains to be established.

\begin{table*}[t]
\centering
\small
\setlength{\tabcolsep}{6pt}
\renewcommand{\arraystretch}{0.95}
\caption{Paired bootstrap differences on the 99 hard-user P4G sessions. Intervals use 10,000 resamples. Positive $\Delta$SR and $\Delta$RCSR and negative $\Delta$AT favor \textsc{MASP}.}
\label{tab:p4g_bootstrap}
\begin{tabular}{@{}lccc@{}}
\toprule
Comparison & $\Delta$SR (95\% CI) & $\Delta$AT (95\% CI) & $\Delta$RCSR (95\% CI) \\
\midrule
\textsc{MASP} $-$ Planner-only & $0.1717\ [0.0505,\ 0.2929]$ & $-0.9899\ [-1.6465,\ -0.3232]$ & $0.1818\ [0.0404,\ 0.3232]$ \\
\textsc{MASP} $-$ SFT & $0.2121\ [0.0909,\ 0.3333]$ & $-1.6768\ [-2.2121,\ -1.1212]$ & $0.1111\ [-0.0202,\ 0.2424]$ \\
\bottomrule
\end{tabular}
\end{table*}

\subsection{P4G Failure Analysis}
\label{app:p4g_failure_analysis}

We analyzed all 21 failures among the 99 Qwen3-4B hard-user P4G dialogues. Failed dialogues end with substantially lower receptivity than successful dialogues ($0.193$ versus $0.713$ on average), and 20 of the 21 failures contain at least three \texttt{refused} labels. Table~\ref{tab:p4g_failure_patterns} summarizes the recurring patterns; categories overlap because one dialogue can exhibit more than one failure mode.

\begin{table}[t]
\centering
\small
\setlength{\tabcolsep}{5pt}
\renewcommand{\arraystretch}{0.95}
\caption{Failure patterns among 21 hard-user P4G failures. Categories are not mutually exclusive.}
\label{tab:p4g_failure_patterns}
\begin{tabular}{@{}lr@{}}
\toprule
Failure pattern & Count \\
\midrule
Financial constraints & $8/21$ \\
Unresolved trust or information needs & $5/21$ \\
Repeated pressure & $10/21$ \\
Conversation breakdowns & $12/21$ \\
\bottomrule
\end{tabular}
\end{table}

The qualitative traces show two common policy errors. In some dialogues, the system continues asking after a clear financial refusal, which increases pressure without changing the user's underlying constraint. In others, it returns to a donation request before resolving trust, transparency, or information needs. These cases suggest that low receptivity should trigger either a more informative response or a respectful termination decision rather than repeated closing attempts.

\subsection{Scope of the Contribution}
\label{app:contribution_scope}

The contribution is scoped to BDI-based mentalized adversarial self play that jointly trains system and user policies for proactive multi-turn dialogue. Prior work has already used self play for dialogue planning and policy learning. The distinction in \textsc{MASP} is the use of an evolving BDI state as the shared structure for teacher--student mentalization, user-relative progress, rationality feedback, and alternating system--user optimization. Within this integrated design, BDI mentalization addresses hidden user states, the rationality signal discourages blind resistance, and alternating self play exposes the system to an adaptive rather than fixed user policy. The ablations in Table~\ref{tab:p4g_ablation} evaluate parts of this pipeline, although comparisons with simpler unstructured latent-state alternatives remain useful future work. The current benchmarks span persuasion, emotional support, empathy, and negotiation, including cooperative and competitive settings, but they do not establish generalization to unseen dialogue domains; broader cross-domain evaluation remains future work.

\section{Additional Method Details}
\label{app:method_details}

\subsection{Training Details}
\label{app:training_details}

Training proceeds in three stages. First, we obtain silver BDI annotations for dialogue prefixes and use them to train a teacher mentalizer and a student mentalizer. The teacher is frozen after pretraining and defines the state used by reward computation, while the student provides the estimate used by the system policy. Second, we warm start both agents with supervised learning on human dialogues, yielding system and user reference policies that initialize self play and serve as KL anchors. We denote the resulting system and user policy adapters as \texttt{pi\_S} and \texttt{pi\_U}, corresponding to $\pi_\theta$ and $\pi_\psi$, respectively.

During alternating adversarial self play, each outer iteration first freezes the user and updates the system, then freezes the system and updates the user. The reported experiments use a GRPO style update without a value function: rollout steps are partitioned into small groups, returns are normalized within each group to form relative advantages, and each LoRA policy is updated with a clipped ratio objective plus KL regularization to its supervised reference. After each self play cycle, the student mentalizer is refreshed on collected prefixes with targets from the frozen teacher. All quantities used for reward computation, including the teacher state, progress signal, and rationality signal, are treated as stop gradient signals during policy optimization.

In implementation, \textsc{MASP} does not modify the causal LM backbone. The student estimate on the system side is rendered into a short BDI control block by retrieving the closest textual BDI record from the BDI prior, while the user side receives the current BDI state. Thus, mentalization enters policy learning through prompts for each role and parameter efficient LoRA updates.

\paragraph{Silver BDI annotations.}
The silver BDI labels are produced by an offline LLM based Oracle Belief Updater (OBU), not by human annotators. For each dialogue prefix, the OBU receives the task description, role mapping, dialogue history, and task specific calibration hints, and is prompted to output exactly six fields: belief, desire, intention, receptivity, confidence, and valence. We use GPT-4o-2024-08-06 for this offline cache construction, with temperature $0.2$ and a maximum of 320 tokens. The resulting cache provides silver supervision for Phase~0 mentalization and the initial states for the BDI prior; reward-side latent state estimates during self play are supplied by the frozen teacher mentalizer.

\paragraph{Concrete BDI record and encoder inputs.}
The following simplified P4G record illustrates the six fields used by the mentalizers.

\begin{promptcard}{Simplified P4G BDI record}
Belief: The user believes that Save the Children is trustworthy and that small donations can help.\\
Desire: The user wants to help children.\\
Intention: The user intends to donate now.\\
Receptivity: $1.0$\\
Confidence: $0.8$\\
Valence: $0.5$
\end{promptcard}

For the teacher mentalizer, the encoder input contains the task description, role mapping, dialogue history, task goal, and user profile when available. The student input contains only the observable dialogue history. In both cases, the predicted belief, desire, and intention strings are encoded separately by the frozen Qwen3 backbone, mean pooled, and $\ell_2$ normalized. The receptivity, confidence, and valence values are then appended as the pre-scaled scalar coordinates in Eq.~(8).

\begin{table}[t]
\centering
\small
\setlength{\tabcolsep}{3.0pt}
\caption{Common hyperparameters used across the completed experiments.}
\label{tab:common_hyperparams}
\begin{tabular}{@{}p{0.52\linewidth}p{0.42\linewidth}@{}}
\toprule
Quantity & Value \\
\midrule
BDI scalar weights $(\alpha_\rho,\alpha_c,\alpha_v)$ & $(1.0,1.0,1.0)$ \\
Discount $\gamma$ & $0.97$ \\
Progress stabilizer $\varepsilon_{\mathrm{prog}}$ & $10^{-6}$ \\
Encoder max length & $384$ \\
Mentalizer projection hidden size / dropout & $512$ / $0.1$ \\
Phase~0 teacher/student learning rate & $3\times10^{-4}$ \\
Phase~0 loss weights $(\lambda_{\mathrm{reg}},\lambda_{\mathrm{dyn}})$ & $(1.0,0.5)$ \\
Phase~1 LoRA learning rate / epochs & $10^{-4}$ / $3$ \\
LoRA rank / alpha / dropout & $64$ / $128$ / $0.05$ \\
LoRA target modules & $q,k,v,o,\mathrm{gate},\mathrm{up},\mathrm{down}$ projections \\
GRPO clip / epochs / max grad norm & $0.2$ / $2$ / $1.0$ \\
Student refresh learning rate / updates & $3\times10^{-4}$ / $12$ \\
\bottomrule
\end{tabular}
\end{table}

\begin{table*}[t]
\centering
\small
\setlength{\tabcolsep}{4.0pt}
\caption{Phase~2 self-play hyperparameters for the completed reported runs. $K$ denotes outer alternating self-play iterations and $\beta$ is the KL coefficient for both policies unless two values are shown.}
\label{tab:phase2_hyperparams}
\begin{tabular}{@{}lcccccc@{}}
\toprule
Run & $K$ & Episodes / iter. & LR & $\beta$ & $(w_{\mathrm{prog}},w_{\mathrm{inf}},w_{\mathrm{rat}})$ & Group size \\
\midrule
P4G, Qwen3-4B & $100$ & $80$ & $10^{-5}$ & $0.05$ & $(1.0,0.25,0.5)$ & $8$ \\
P4G, Qwen3-0.6B & $100$ & $80$ & $2\times10^{-6}$ & $0.10$ & $(1.0,0.25,0.5)$ & $8$ \\
ESConv, Qwen3-4B & $12$ & $64$ & $2\times10^{-6}$ & $0.08$ & $(1.0,0.25,0.0)$ & $8$ \\
ESConv, Qwen3-0.6B & $40$ & $64$ & $2\times10^{-6}$ & $0.08$ & $(1.0,0.25,0.0)$ & $8$ \\
EmpatheticDialogues, both backbones & $40$ & $64$ & $2\times10^{-6}$ & $0.08$ & $(1.0,0.25,0.0)$ & $8$ \\
CraigslistBargain, both backbones & $40$ & $64$ & $2\times10^{-6}$ & $0.08$ & $(1.0,0.25,0.0)$ & $8$ \\
\bottomrule
\end{tabular}
\end{table*}

\subsection{Target States and BDI Prior}
\label{app:target_states}

We use one fixed target-state template per task. Each template is specified before evaluation and shared across all examples in that task; it is not tuned per test dialogue. In compressed form, the templates are:
\begin{itemize}
    \item \textbf{P4G}: the user believes Save the Children is trustworthy and that small donations can help, wants to help children, and intends to donate now; $(\rho,c,v)=(1.0,0.8,0.5)$.
    \item \textbf{ESConv}: the user believes talking with the supporter and trying suggestions can help, wants to feel less distressed and find coping strategies, and intends to open up and try a concrete suggestion; $(\rho,c,v)=(1.0,0.7,0.4)$.
    \item \textbf{EmpatheticDialogues}: the user believes the listener understands their feeling, wants to feel heard and emotionally connected, and intends to continue sharing openly; $(\rho,c,v)=(0.8,0.6,0.3)$.
    \item \textbf{CraigslistBargain}: the seller believes the buyer's offer can be mutually acceptable, wants to close at an acceptable price, and intends to agree on a final transaction; $(\rho,c,v)=(0.8,0.7,0.3)$.
\end{itemize}
The BDI prior used for initialization and prompt rendering is the non-parametric pool of initial BDI records extracted from the training split. During prompt conditioning, a continuous estimate $\hat z_t$ is mapped back to text by cosine nearest-neighbor retrieval over this encoded pool, and the retrieved BDI record is rendered as the system-side control block.

\subsection{Auxiliary Training Objectives}
\label{app:aux_losses}

Let $\bar z_t$ denote the silver latent state annotation produced by the offline OBU. We train the teacher mentalizer with
\begin{equation}
\begin{aligned}
\mathcal{L}_F
&=
\lambda_{\mathrm{reg}}\sum_t \|F_\omega(h_t,p_u,g)-\bar z_t\|_A^2 \\
&\quad+
\lambda_{\mathrm{dyn}}\sum_t \|\Delta z_t^F-\Delta \bar z_t\|_A^2,
\end{aligned}
\end{equation}
where
\begin{equation}
\begin{aligned}
\Delta z_t^F &= F_\omega(h_{t+1},p_u,g)-F_\omega(h_t,p_u,g), \\
\Delta \bar z_t &= \bar z_{t+1}-\bar z_t.
\end{aligned}
\end{equation}

The student mentalizer is trained to mimic the frozen teacher:
\begin{equation}
\begin{aligned}
\mathcal{L}_M
&=
\sum_t
\left\|
M_\phi(h_t)-\mathrm{sg}(F_\omega(h_t,p_u,g))
\right\|_A^2,
\end{aligned}
\end{equation}
where $\mathrm{sg}(\cdot)$ denotes stop gradient. During self play, we continue to refresh the student with teacher targets collected from self play prefixes and mix in real dialogue prefixes from the BDI cache, which prevents the student from drifting purely toward the current self play distribution.

The rationality signal is supplied by a fixed judge:
\begin{equation}
q_t=R_{\mathrm{rat}}(h_t,z_t,a_t,u_t)\in\{-1,0,1\}.
\end{equation}
The judge returns a positive signal when the user response is rational under the current mental state, a negative signal when it is irrational, and zero when the output cannot be parsed. This signal is used as reward feedback and for RCSR evaluation, but it is not differentiated through during policy optimization.

\subsection{Policy Optimization}
\label{app:policy_opt}

The main experiments optimize both agents with GRPO at the utterance level. For each role, we collect rollout steps and partition them into groups $\mathcal{G}$. Let $G_i$ be the discounted return for response $y_i$ in group $\mathcal{G}$, and define the advantage relative to the group
\begin{equation}
\hat A_i=
\frac{G_i-\mathrm{mean}_{j\in\mathcal{G}}G_j}
{\mathrm{std}_{j\in\mathcal{G}}G_j+\epsilon}.
\end{equation}
For a generic policy $\pi_\eta$, the clipped objective at the response level is
\begin{equation}
\begin{aligned}
\mathcal{L}_{\mathrm{GRPO}}(\eta)
&=
-\mathbb{E}_i\Big[
\min\big(
r_i(\eta)\hat A_i,\\
&\qquad
\mathrm{clip}\big(
r_i(\eta),1-\epsilon,1+\epsilon
\big)\hat A_i
\big)
\Big] \\
&\quad+
\beta\,\mathrm{KL}\big(
\pi_\eta(\cdot\mid o_i)\,\|\\
&\qquad
\pi_{\mathrm{ref}}(\cdot\mid o_i)
\big),
\end{aligned}
\end{equation}
where
\begin{equation}
\begin{aligned}
r_i(\eta)
&=
\exp\!\Big(
\log \pi_\eta(y_i\mid o_i)\\
&\qquad
- \log \pi_{\eta_{\mathrm{old}}}(y_i\mid o_i)
\Big).
\end{aligned}
\end{equation}
For the system, $y_i=a_i$ and $o_i=(h_i,\hat z_i,g)$; for the user, $y_i=u_i$ and $o_i=(h_i,z_i,p_u,g)$. The reference policy $\pi_{\mathrm{ref}}$ is the corresponding supervised warm start policy.

\subsection{Prompt Conditioning}
\label{app:conditioning}

We do not modify the backbone architecture. Instead, the latent state is rendered into a short text block and prepended to the prompt for each role. The system consumes the student estimate $\hat z_t$ through cosine nearest-neighbor retrieval over the BDI prior defined in Appendix~\ref{app:target_states}, while the user policy is prompted with the current BDI state $z_t$. This keeps the framework compatible with standard causal language models and LoRA based training.

\section{Dataset and Evaluation Details}
\label{app:dataset_details}
\label{app:eval_details}

\subsection{Dataset Statistics}
\label{app:dataset_statistics}

Table~\ref{tab:dataset_stats} provides dataset statistics and role mappings for the benchmark settings used or prepared in this work. 
\emph{Corpus Size} reports the size of the released dataset, while \emph{Eval Size} denotes the number of held out dialogue contexts used to construct soft user and hard user rollouts when that benchmark is evaluated. 
\begin{table*}[t]
\centering
\small
\setlength{\tabcolsep}{6pt}
\renewcommand{\arraystretch}{1.05}
\begin{tabular*}{\textwidth}{@{\extracolsep{\fill}}lllll@{}}
\toprule
Dataset & Task & Roles & Corpus Size & Eval Size \\
\midrule
P4G & Persuasion & Persuader / Persuadee & 1,017 dialogues & 99 \\
ESConv & Emotional support & Supporter / Seeker & 1,300 dialogues & 195 \\
EmpatheticDialogues & Empathetic response & Speaker / Listener & 24,850 dialogues & 2,542 \\
CraigslistBargain & Negotiation & Buyer / Seller & 6,682 dialogues & 838 \\
\bottomrule
\end{tabular*}
\caption{
Dataset statistics. Eval size denotes the number of held out dialogue contexts prepared for closed loop evaluation.
}
\label{tab:dataset_stats}
\end{table*}

\subsection{Common Closed Loop Evaluation Protocol}
\label{app:common_eval_protocol}

All benchmarks are evaluated as closed loop rollouts with an eight turn budget. At each system turn, the system policy generates a response conditioned on the dialogue history and the available mental state estimate. The user side is either the soft imitation style user policy or the hard adversarial user policy. After each exchange between assistant and user, a task specific success judge assigns the outcome label used to decide whether the task specific success criterion has been met. Unless otherwise noted, the success judge uses GPT-3.5 with three sampled judgments. We use the following common rationality judge prompt.

\begin{promptcard}{Common rationality judge}
Task context: \textit{[task description]}\\
User mental state: \textit{[current BDI state]}\\
Most recent exchange:\\
Assistant: \textit{[assistant turn]}\\
User: \textit{[user turn]}\\[0.35em]
Question: Was it rational for a user with that internal mental state to respond this way? A rational response means that the user updates when the assistant gives a concrete and honest reason to update, or reasonably resists when such a reason is absent.\\[0.35em]
Answer with exactly one label: \texttt{RATIONAL} or \texttt{NOT\_RATIONAL}.
\end{promptcard}

Each episode begins from a held-out initial context $h_0$. At turn $t$, the system produces $a_t$ from the complete current history and its estimated user state, the user simulator produces $u_t$, and both utterances are appended to form $h_{t+1}$. The success judge then reads the complete accumulated dialogue through $h_{t+1}$. If the goal has not been reached, the next exchange continues from that same history, for at most eight system--user exchanges. The judge is therefore a per-turn checkpoint over a growing trajectory rather than an evaluator of an isolated response.

Table~\ref{tab:p4g_rollout_example} shows an actual successful P4G rollout. The system first elicits the user's stance, then provides task-relevant information, and only then makes a concrete request after the user becomes more receptive.

\begin{table*}[t]
\centering
\small
\setlength{\tabcolsep}{5pt}
\renewcommand{\arraystretch}{1.05}
\caption{Example closed-loop P4G trajectory with the success-judge checkpoint after each accumulated exchange.}
\label{tab:p4g_rollout_example}
\begin{tabular}{@{}cp{0.36\textwidth}p{0.36\textwidth}l@{}}
\toprule
Turn & System & User & Judge \\
\midrule
1 & Have you heard of Save the Children? & No, tell me about it. & Neutral; unsuccessful \\
2 & It promotes children's rights and supports children in need. & That sounds worthwhile. & Positive; unsuccessful \\
3 & Would you donate \$0.25 from your task payment? & I'm happy to donate today. & Agree; successful \\
\bottomrule
\end{tabular}
\end{table*}

\subsection{P4G Evaluation Protocol}
\label{app:p4g_eval_protocol}

P4G success means that the persuadee has decided to donate to Save the Children. The judge maps each sampled judgment to one of four labels, with rewards \texttt{refused} $=-1.0$, \texttt{neutral} $=-0.5$, \texttt{positive} $=0.1$, and \texttt{agree} $=1.0$. The reported P4G rows use a success threshold of $0.6$ over the averaged sampled reward. This threshold is fixed before evaluation and makes binary success closer to an \texttt{agree} decision than a merely positive response; with three sampled judgments, for example, two \texttt{agree} labels plus one \texttt{positive} label pass, while two \texttt{agree} labels plus a \texttt{neutral} or \texttt{refused} label do not. The complete \textsc{MASP} pipeline uses a Qwen3 backbone with Phase 0 teacher and student mentalization, Phase 1 supervised warm starts for \texttt{pi\_S} and \texttt{pi\_U}, and Phase 2 alternating GRPO self play. The reported Qwen3-4B P4G rows use the Phase 2 system adapter together with the P4G scaffold used at evaluation time: \texttt{p4g\_warmup\_v1} and adaptive dynamic close injection. For system decoding, \textsc{MASP} and Planner-only use temperature $0.3$, while SFT uses temperature $0.75$; the user simulator uses temperature $0.9$. We use the 99 example clean test split.

\begin{promptcard}{P4G success judge}
Conversation so far:\\
\textit{[dialogue history]}\\
Assistant: \textit{[assistant action]}\\
Persuadee: \textit{[user reply]}\\[0.35em]
Question: Has the persuadee been successfully convinced to make a donation at the end of the conversation?\\[0.35em]
Choose exactly one response: the persuadee has explicitly refused; the persuadee remains neutral about donating; the persuadee has a positive attitude towards donating but has not decided yet; the persuadee has decided to donate.
\end{promptcard}

\subsection{Human--AI Agreement Audit}
\label{app:human_ai_agreement}

Following the dialogue-level evaluation practice used in prior proactive dialogue work \citep{Deng2023PlugandPlayPP,Rakib2025DialogXpertDI}, we conducted a supplementary human--AI agreement audit for the P4G success judge. We randomly sampled 100 P4G evaluation records from the union of the validation and test splits and asked five human annotators to label the final outcome, binary success, success or terminal turn, and turn-level task labels. The same records were evaluated by the automatic GPT-3.5 judge used in our closed-loop experiments. This audit is intended as a sanity check for the automatic judge rather than as a standalone human evaluation of policy quality. On the coarse endpoint labels, we observed no disagreements between human and GPT-3.5 annotations: final outcome and binary success both reached 100.0\% exact agreement, corresponding to Cohen's $\kappa=1.000$. To avoid relying only on these endpoint labels, we also report finer-grained timing and turn-level agreement. For successful cases, the success-turn exact match was 97.4\% with a mean absolute error of 0.039 turns; over all records, terminal-turn exact match was 98.0\% with a mean absolute error of 0.030 turns. Turn-level task labels reached 95.7\% agreement with Cohen's $\kappa=0.942$. These non-endpoint metrics provide a more conservative view of judge reliability and show that the remaining disagreements mainly concern success timing or intermediate task states rather than the final donation decision.

\subsection{ESConv Evaluation Protocol}
\label{app:esconv_eval_protocol}

For ESConv, the automatic judge estimates whether the support interaction improves the help seeker's emotional state. The judge uses the DialogXpert/PPDPP style four level emotional outcome labels with rewards \texttt{worse} $=-1.0$, \texttt{same} $=-0.5$, \texttt{somewhat\_better} $=0.5$, and \texttt{resolved} $=1.0$. Following that convention, the reported ESConv SR treats the averaged sampled reward as successful when it is greater than or equal to $0.5$, so both \texttt{somewhat\_better} and \texttt{resolved} count as successful. This criterion measures whether the user appears meaningfully helped, not whether the underlying emotional issue is fully resolved; \texttt{resolved} is therefore a stricter diagnostic outcome rather than the primary SR threshold. For system decoding, \textsc{MASP} and Planner-only use temperature $0.3$, while SFT uses temperature $0.75$; the user simulator uses temperature $0.9$. For the Qwen3-0.6B hard user \textsc{MASP} run, 194 of 195 first successful turns are labeled \texttt{somewhat\_better}, while only one is labeled \texttt{resolved}, confirming that the near ceiling SR mainly reflects rapid improvement rather than complete resolution.

\begin{promptcard}{ESConv success judge}
Conversation between a Therapist and a Patient about \textit{[emotion type]} regarding \textit{[problem type]}:\\
\textit{[dialogue history]}\\
Therapist: \textit{[assistant action]}\\
Patient: \textit{[user reply]}\\[0.35em]
Question: Has the Patient's issue been solved?\\[0.35em]
Choose exactly one response: No, the Patient feels worse; No, the Patient feels the same; No, but the Patient feels better; Yes, the Patient's issue has been solved.
\end{promptcard}

\subsection{EmpatheticDialogues Evaluation Protocol}
\label{app:empathetic_eval_protocol}

EmpatheticDialogues is originally a next response empathy benchmark rather than a goal completion benchmark, so we use a closed loop success criterion based on emotional resolution. Each trajectory is judged with the labels \texttt{worse}, \texttt{same}, \texttt{somewhat\_better}, and \texttt{resolved}. For the reported EmpatheticDialogues rows, success requires the final outcome label to be \texttt{resolved}; \texttt{somewhat\_better} is retained as a diagnostic outcome but is not counted as task success. For system decoding, \textsc{MASP} and Planner-only use temperature $0.3$, while SFT uses temperature $0.75$; the user simulator uses temperature $0.9$. All rows use the 2,542 example test split and max eight listener turns.

\begin{promptcard}{EmpatheticDialogues success judge}
Conversation so far:\\
\textit{[dialogue history]}\\
Listener: \textit{[assistant action]}\\
Speaker: \textit{[user reply]}\\[0.35em]
Question: Has the speaker's emotional state improved or been resolved? For this dataset, \texttt{resolved} means the speaker shows clear evidence of feeling accurately heard, validated, emotionally steadier, and willing to continue. A generic sympathetic response is at most \texttt{somewhat\_better}.\\[0.35em]
Respond with one exact label only: \texttt{worse}, \texttt{same}, \texttt{somewhat\_better}, or \texttt{resolved}.
\end{promptcard}

\subsection{CraigslistBargain Evaluation Protocol}
\label{app:craigslist_eval_protocol}

CraigslistBargain success means that the buyer and seller reach an agreement that satisfies the utility criterion for the buyer. The judge first determines whether a deal has been reached and extracts the final agreed price. If any sampled judgment indicates no deal, the outcome is treated as \texttt{no\_deal}. Otherwise, the final deal price is converted to buyer normalized utility, $(p_{\mathrm{deal}}-p_{\mathrm{seller}})/(p_{\mathrm{buyer}}-p_{\mathrm{seller}})$, and the interaction is successful when this value meets the success threshold. In our CraigslistBargain evaluation, this threshold is fixed to $0.0$, so SR@8 counts deals that close within eight turns and achieve nonnegative buyer utility. The main table therefore uses the same paper-level metrics as the other datasets: SR@8, AT, and RCSR@8. Deal completion without the buyer utility threshold is treated only as an auxiliary diagnostic outside the main metric set. For this dataset, the system role is the buyer and the user role is the seller. We therefore define hard and soft by buyer-side difficulty rather than by the training-stage label: the Phase 1 BC seller is reported as the hard user because it is more difficult for the buyer-system to close favorable deals, while the Phase 2 self-play seller is reported as the soft user because it accepts buyer-favorable deals more often in our evaluation. This reversal is specific to buyer-system CraigslistBargain and does not change the soft/hard convention for the cooperative P4G, ESConv, or EmpatheticDialogues tasks.

\begin{promptcard}{CraigslistBargain success judge}
Conversation so far:\\
\textit{[dialogue history]}\\
Buyer: \textit{[assistant action]}\\
Seller: \textit{[user reply]}\\[0.35em]
Question: Have the buyer and seller successfully reached a deal at the end of the conversation?\\[0.35em]
If they have reached a deal, reply exactly: \texttt{They have reached a deal at [price].} Otherwise, reply exactly: \texttt{They have not reached a deal.}
\end{promptcard}
\end{document}